%% file: litreviews.tex
\documentclass[11pt,a4paper,dvipsnames,table]{article}
\usepackage{authblk}
\usepackage[hyperref]{emnlp2021}
\usepackage{times}
\usepackage{latexsym}
\usepackage[T1]{fontenc}
\usepackage[utf8]{inputenc}
\usepackage{microtype}

\usepackage{enumitem}
\usepackage{graphicx}
\usepackage{xspace}
\usepackage{booktabs}
\usepackage{tabularx}
\usepackage{amsmath}
\usepackage{pdflscape}

\usepackage{hyphenat}
\usepackage{soul}
\usepackage{multirow}
\usepackage{colour-blind}

\usepackage{microtype}

\input{commands.tex}

\title{\litreview: A Dataset for Multi-Document Summarization of Medical Studies}

\author{\textbf{Jay DeYoung}$^{1}$\thanks{\enspace Work performed during internship at AI2}\quad \textbf{Iz Beltagy}$^{2}$\quad \textbf{Madeleine van Zuylen}$^{2}$\quad \textbf{Bailey Kuehl}$^{2}$\quad \textbf{Lucy Lu Wang}$^{2}$\\
Northeastern University$^{1}$ \quad Allen Institute for AI$^{2}$ \\
\texttt{deyoung.j@northeastern.edu}\\
\texttt{\{beltagy,madeleinev,baileyk,lucyw\}@allenai.org}
}

\date{}

\begin{document}
\maketitle

\begin{abstract}
To assess the effectiveness of any medical intervention, researchers must conduct a time-intensive and manual literature review. NLP systems can help to automate or assist in parts of this expensive process. In support of this goal, we release \litreview (\textbf{M}ulti-Document \textbf{S}ummarization of \textbf{M}edical \textbf{S}tudies), a dataset of over \numstudies documents and \numreviews summaries derived from the scientific literature. This dataset facilitates the development of systems that can assess and aggregate contradictory evidence across multiple studies, and is the first large-scale, publicly available multi-document summarization dataset in the biomedical domain. 
We experiment with a summarization system based on BART, with promising early results, though significant work remains to achieve higher summarization quality.
We formulate our summarization inputs and targets in both free text and structured forms and modify a recently proposed metric to assess the quality of our system's generated summaries. Data and models are available at \githublink.
\end{abstract}

\input{sections/introduction.tex}

\input{sections/background.tex}

\input{sections/dataset}

\input{sections/models}

\input{sections/related_work}

\input{sections/discussion}

\input{sections/conclusions.tex}

\section*{Acknowledgements}
This project is supported in part by NSF Grant OIA-2033558. We thank Ani Nenkova, Byron Wallace, Dan Weld, the reviewers, and members of the Semantic Scholar team for their valuable feedback.


\input{sections/broader_impact}

\bibliography{litreview}
\bibliographystyle{acl_natbib}

\clearpage
\input{sections/appendix.tex}

\end{document}

%% file: commands.tex
\newcommand{\pop}[1]{{\textit{\textcolor{cb-blue-sky}{#1}}}}
\newcommand{\inte}[1]{{\textcolor{cb-green-lime2}{\underline{#1}}}}
\newcommand{\out}[1]{{\textbf{\textcolor{cb-rose}{#1}}}}

\newcommand\litreview{{MS\^{}2}\xspace}

\newcommand\semanticscholar{Semantic Scholar\xspace}
\newcommand\githublink{\url{https://github.com/allenai/ms2}\xspace}

\newcommand\texttotext{\textit{texts-to-text}\xspace}
\newcommand\tabtotab{\textit{table-to-table}\xspace}
\newcommand\background{\textsc{background}\xspace}
\newcommand\target{\textsc{target}\xspace}
\newcommand\other{\textsc{other}\xspace}
\newcommand\increases{\textit{increases}\xspace}
\newcommand\decreases{\textit{decreases}\xspace}
\newcommand\nochange{\mbox{\textit{no\_change}}\xspace}
\newcommand\longbart{LED\xspace}
\newcommand\standard{BART\xspace}

\newcommand\numreviews{20K\xspace}
\newcommand\numstudies{470k\xspace} 
\newcommand\numtestset{2K\xspace}
\newcommand\numsentannots{3000\xspace} 
\newcommand\numsentarticles{220\xspace}
\newcommand\jsdmetric{$\Delta$EI\xspace}
\newcommand\MDS{MDS\xspace}

\newcolumntype{L}[1]{>{\raggedright\let\newline\\\arraybackslash\hspace{0pt}}p{#1}}

%% file: sections/introduction.tex
\section{Introduction}
\label{sec:intro}

Multi-document summarization (MDS) is a challenging task, with relatively limited resources and modeling techniques. 
Existing datasets are either in the general domain, such as WikiSum \citep{2018generating} and Multi-News \citep{multinews}, 
or very small such as DUC\footnote{\label{footnote:duc}\href{https://duc.nist.gov}{https://duc.nist.gov}} or TAC 2011 \citep{tac2011}. 
In this work, we add to this burgeoning area by developing a dataset for summarizing biomedical findings. 
We derive documents and summaries from systematic literature reviews, a type of biomedical paper that synthesizes results across many other studies.
Our aim in introducing \litreview is to: (1) expand \MDS to the biomedical domain, (2) investigate fundamentally challenging issues in NLP over scientific text, such as summarization over contradictory information and assessing summary quality via a structured intermediate form, and (3) aid in distilling large amounts of biomedical literature by supporting automated generation of literature review summaries.

\begin{figure}[t!]
    \centering
    \includegraphics[width=\linewidth]{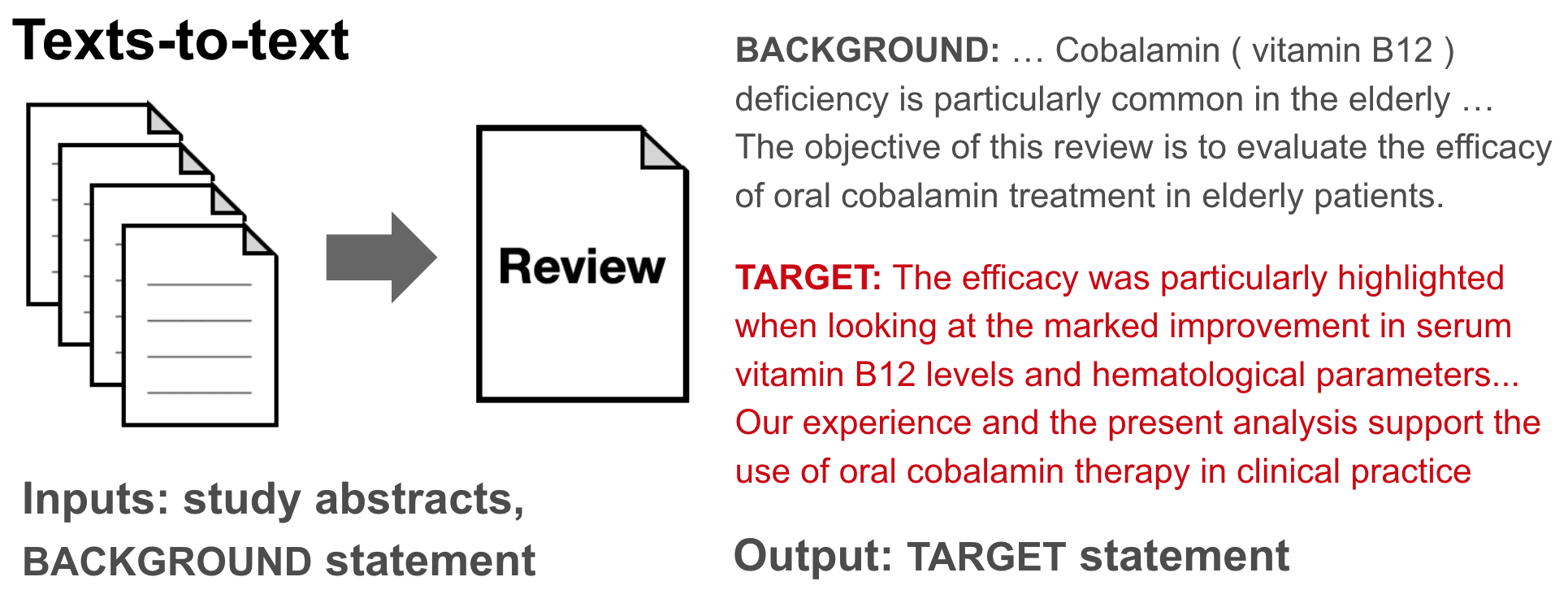}
    \caption{Our primary formulation (\texttotext) is a \textit{seq2seq} \MDS task. Given study abstracts and a \background statement, generate the \target summary.}
    \label{fig:tasks}
    \vspace{-0.2em}
\end{figure}

\begin{figure}[tbp!]
    \centering
    \includegraphics[width=0.98\linewidth]{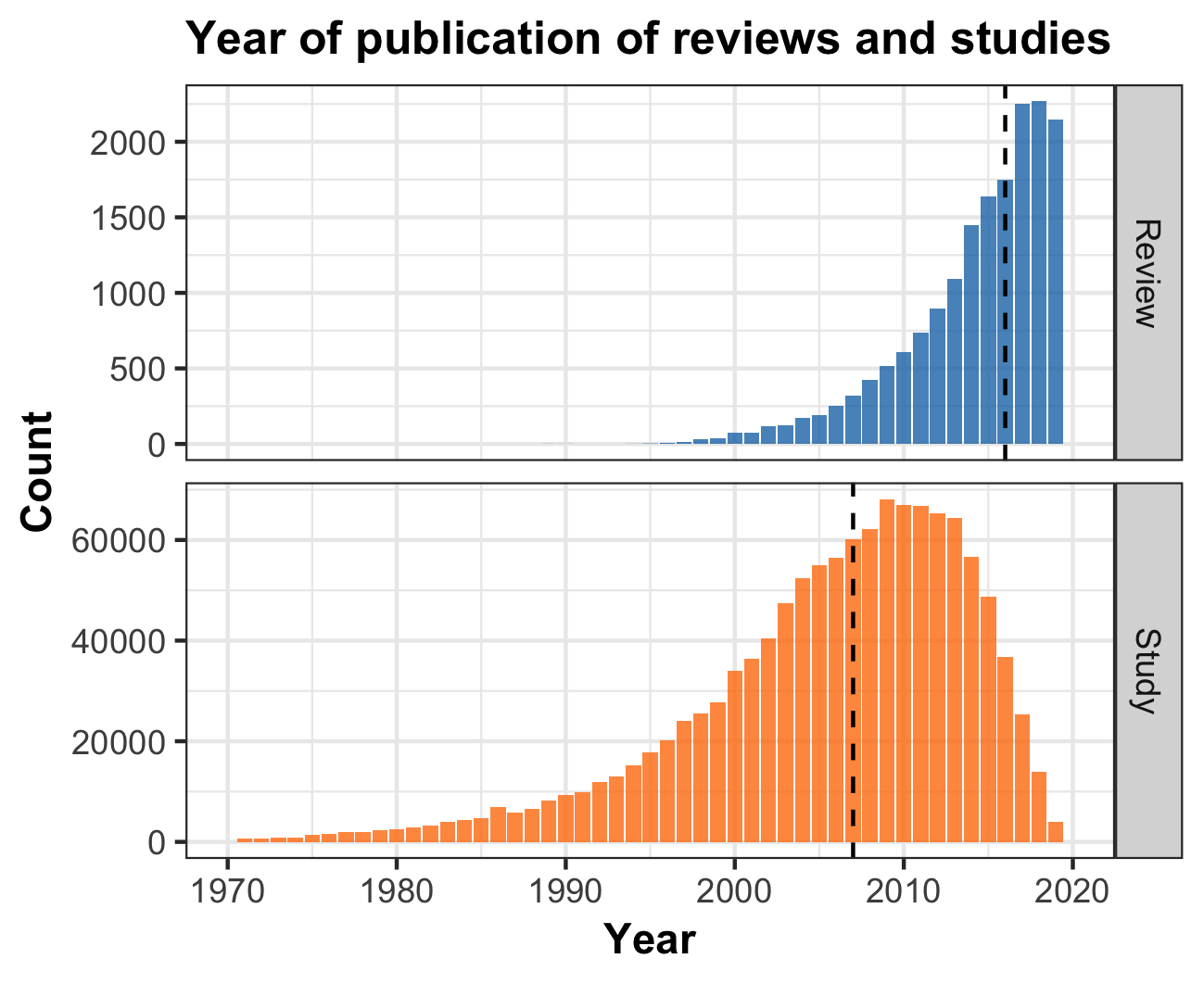}
    \caption{The distributions of review and study publication years in \litreview show a clear temporal lag. Dashed lines mark the median year of publication.}
    \label{fig:delay}
    \vspace{-1.0em}
\end{figure}

\emph{Systematic reviews} synthesize  knowledge across many studies \citep{Khan2003FiveST}, and they are so called for the systematic (and expensive) process of creating a review; each taking 1-2 years to complete \citep{Michelson2019TheSC}.\footnote{\href{https://community.cochrane.org/review-production/production-resources/proposing-and-registering-new-cochrane-reviews}{https://community.cochrane.org/review-production/ production-resources/proposing-and-registering-new-cochrane-reviews}} 
As we note in Fig.~\ref{fig:delay}, a delay of around 8 years is observed between reviews and the studies they cite!
The time and cost of creating and updating reviews has inspired efforts at automation \citep{Tsafnat2014SystematicRA,Marshall2016RobotReviewerEO,Beller2018MakingPW,Marshall2019TowardSR}, and the constant deluge of studies\footnote{Of the heterogeneous study types, randomized control trials (RCTs) offer the highest quality of evidence. Around 120 RCTs are published per day as of this writing \href{https://ijmarshall.github.io/sote/}{https://ijmarshall.github.io/sote/}, up from 75 in 2010 \citep{75trials}.} has only increased this need. 

To move the needle on these challenges and support further work on literature review automation, we present \litreview, a multi-document summarization dataset in the biomedical domain. 
Our contributions in this paper are as follows:
\begin{itemize}[noitemsep,leftmargin=*]
    \item We introduce \litreview, a dataset of \numreviews reviews and \numstudies studies summarized by these reviews.
    \item We define a \texttotext \MDS task (Fig.~\ref{fig:tasks}) based on \litreview, by identifying target summaries in each review and using study abstracts as input documents. We develop a BART-based model for this task, which produces fluent summaries that agree with the evidence direction stated in gold summaries around 50\% of the time.  
    \item In order to expose more granular representations to users, we define a structured form of our data to support a \tabtotab task (\S~\ref{sec:table2table}). 
    We leverage existing biomedical information extraction systems \citep{ebmnlp,evidenceinference} (\S \ref{sec:pico_tagging}, \S \ref{sec:evidence_inference}) to evaluate agreement between target and generated summaries.
\end{itemize}

%% file: sections/background.tex
\section{Background}
\label{sec:background}

Systematic reviews aim to synthesize results over all relevant studies on a topic, providing high quality evidence for biomedical and public health decisions. They are a fixture in the biomedical literature, with many established protocol around their registration, production, and publication \cite{Chalmers2002ABH, Starr2009TheOE, Booth2012TheNA}. Each systematic review addresses one or several research questions, and results are extracted from relevant studies and summarized. 
For example, a review investigating the effectiveness of Vitamin B12 supplementation in older adults \citep{Andrs2010EfficacyOO} synthesizes results from 9 studies.

The research questions in systematic reviews can be described using the PICO framework \citep{Zakowski2004EvidencebasedMA}. PICO (which stands for \underline{P}opulation: \textit{who is studied?} \underline{I}ntervention: \textit{what intervention was studied?} \underline{C}omparator: \textit{what was the intervention compared against?} \underline{O}utcome: \textit{what was measured?}) 
 defines the main facets of biomedical research questions, and 
allows the person(s) conducting a review to identify relevant studies (studies included in a review generally have the same or similar PICO elements as the review). 
A medical systematic review is one which reports results for applying any kind of medical or social intervention to a group of people. Interventions are wide-ranging, including yoga, vaccination, team training, education, vitamins, mobile reminders, and more. 
Recent work on evidence inference \citep{evidenceinference,nye2020understanding} goes beyond identifying PICO elements, and aims to group and identify overall findings in reviews.
\litreview is a natural extension of these paths: we create a dataset and build a system with both natural summarization targets from input studies, while also incorporating the inherent structure studied in previous work.

In this work, we use the term \textit{review} when describing literature review papers, which provide our summary targets. We use the term \textit{study} to describe the documents that are cited and summarized by each review. There are various study designs which offer differing levels of evidence, e.g. clinical trials, cohort studies, observational studies, case studies, and more \citep{Concato2000RandomizedCT}. Of these study types, randomized controlled trials (RCTs) offer the highest quality of evidence \citep{Meldrum2000ABH}.

%% file: sections/dataset.tex
\section{Dataset}\label{section:dataset}
We construct \litreview from papers in the \semanticscholar literature corpus \citep{Ammar2018ConstructionOT}. 
First, we create a corpus of reviews and studies based on the suitability criteria defined in \S \ref{subsection:suitability_def}. 
For each review, we classify individual sentences in the abstract to identify summarization targets (\S \ref{subsection:sentence_labeling_def}).
We augment all reviews and studies with PICO span labels and evidence inference classes as described in \S \ref{sec:pico_tagging} and \S \ref{sec:evidence_inference}.
As a final step in data preparation, we cluster reviews by topic and form train, development, and test sets from these clusters (\S \ref{subsection:clustering_def}).

\subsection{Identifying suitable reviews and studies} 
\label{subsection:suitability_def}
To identify suitable reviews, we apply (i) a high-recall heuristic keyword filter, (ii) PubMed filter, (iii) study-type filter, and (iv) suitability classifier, in series. 
The keyword filter looks for the phrase ``systematic review'' in the title and abstracts of all papers in \semanticscholar, which yields 220K matches. 
The PubMed filter, yielding 170K matches, limits search results to papers that have been indexed in the PubMed database, which restricts reviews to those in the biomedical, clinical, psychological, and associated domains. We then use citations and Medical Subject Headings (MeSH) to identify input studies via their document types and further refine the remaining reviews, see App.~\ref{appendix:mesh_filtering} for details on the full filtering process. 

Finally, we train a suitability classifier as the final filtering step, using SciBERT \citep{beltagy-etal-2019-scibert}, a BERT \citep{bert} based language model trained on scientific text. Details on classifier training and performance are provided in Appendix~\ref{appendix:suitability_classifier}. Applying this classifier to the remaining reviews leaves us with \numreviews candidate reviews. 

\subsection{Background and target identification}
\label{subsection:sentence_labeling_def}
\begin{table}[tb!]
    \centering
    \scriptsize
    \begin{tabular}{p{15mm}p{54mm}}
        \toprule
        \small{Label} & \small{Sentence} \\
        \midrule
        \background & ... AREAS COVERED IN THIS REVIEW The objective of this review is to evaluate the {\out{efficacy of}} {\inte{oral cobalamin treatment}} in {\pop{elderly patients .}} \\
        \hline
        \other & To reach this objective , PubMed data were systematic ally search ed for {\pop{English and French articles published from January 1990 to July 2008 .} ...} \\
        \hline
        \target & The {\out{efficacy}} was particularly highlighted when looking at the marked improvement in {\out{serum vitamin B12 levels and hematological parameters}} , for example {\out{hemoglobin level , mean erythrocyte cell volume and reticulocyte count .}} \\
        \hline
        \other & The {\out{effect}} of {\inte{oral cobalamin treatment}} in {\pop{patients}} presenting with {\out{severe neurological manifestations}} has not yet been adequately documented .... \\
        \bottomrule
    \end{tabular}
    \caption{Abbreviated example from \citet{Andrs2010EfficacyOO} with predicted sentence labels (full abstract in Tab.~\ref{tab:sent_example_full}, App.~\ref{app:sentence_classification_example}). Spans corresponding to {\pop{Population}}, {\inte{Intervention}}, and {\out{Outcome}} elements are tagged and surrounded with special tokens.}
    \vspace{-1em}
    \label{tab:sent_example}
\end{table}

For each review, we identify two sections: 1) the \background statement, which describes the research question, and 2) the overall effect or findings statement as the \target of the \MDS task (Fig.~\ref{fig:tasks}). We frame this as a sequential sentence classification task \citep{cohan-etal-2019-pretrained}: given the sentences in the review abstract, classify them as \background, \target, or \other. All \background sentences are aggregated and used as input in modeling.
All \target sentences are aggregated and form the summary target for that review. Sentences classified as \other may describe the methods used to conduct the review, detailed findings such as the number of included studies or numerical results, as well as recommendations for practice. \other sentences are not suitable for modeling because they either contain information specific to the review, as in methods; too much detail, in the case of results; or contain guidance on how medicine should be practiced, which is both outside the scope of our task definition and ill-advised to generate.

Five annotators with undergraduate or graduate level biomedical background labeled \numsentannots sentences from \numsentarticles review abstracts. During annotation, we asked annotators to label sentences into 9 classes (which we collapse into the 3 above; see App.~\ref{appendix:evidence_sentence_annotation} for detailed info on other classes). Two annotators then reviewed all annotations and corrected mistakes. The corrections yield a Cohen's $\kappa$ \citep{Cohen1960ACO} of 0.912. 
Though we retain only \background and \target sentences for modeling, we provide labels to all 9 classes in our dataset.

\begin{table}[t!]
\scriptsize
\setlength{\tabcolsep}{2pt}
\centering
    \begin{tabular}
    {L{13mm}L{13mm}L{34mm}p{11mm}}
    \toprule
         Intervention & Outcome & Evidence sentence & Direction \\
    \midrule
         oral cobalamin therapy & effect & The effect of oral cobalamin treatment in patients presenting with severe neurological manifestations has not yet been adequately documented . & \nochange \\
    \midrule
        oral cobalamin therapy & discomfort, inconvenience and cost & Oral cobalamin treatment avoids the discomfort , inconvenience and cost of monthly injections . & \decreases  \\
    \midrule
        oral cobalamin therapy & serum vitamin b12 levels and hematological parameters & The efficacy was particularly highlighted when looking at the marked improvement in serum vitamin B12 levels and hematological parameters , for example hemoglobin level , mean erythrocyte cell volume and reticulocyte count . & \increases \\
    \bottomrule
    \end{tabular}
    \caption{Sample Intervention, Outcome, evidence statement, and identified effect directions from a systematic review investigating the effectiveness of vitamin B12 therapies in the elderly \citep{Andrs2010EfficacyOO}. }
    \label{tab:sample_pico}
    \vspace{-2.5em}
\end{table}

Using SciBERT \citep{beltagy-etal-2019-scibert}, we train a sequential sentence classifier. 
We prepend each sentence with a \texttt{[SEP]} token and use a linear layer followed by a softmax to classify each sentence. 
A detailed breakdown of the classifier scores is available in Tab.~\ref{tab:detailed_sentence_breakdown_scores}, App.~\ref{appendix:evidence_sentence_annotation}. While the classifier performs well (94.1 F1) at identifying \background sentences, it only achieves 77.4 F1 for \target sentences. The most common error for \target sentences is confusing them for results from individual studies or detailed statistical analysis.
Tab.~\ref{tab:sent_example} shows example sentences with predicted labels. 
Due to the size of the dataset, we cannot manually annotate sentence labels for all reviews, so we use the sentence classifier output as silver labels in the training set. 
To ensure the highest degree of accuracy for the summary targets in our test set, we manually review all 4519 \target sentences in the \numtestset reviews of the test set, correcting 1109 sentences.
Any reviews without \target sentences are considered unsuitable and are removed from the final dataset.

\subsection{Structured form}
As discussed in \S\ref{sec:background}, the key findings of studies and reviews can be succinctly captured in a structured representation. The structure consists of PICO elements \citep{ebmnlp} that define what is being studied, in addition to the effectiveness of the intervention as inferred through Evidence Inference (\S\ref{sec:evidence_inference}). In addition to the textual form of our task, we construct this structured form and release it with \litreview to facilitate investigation of consistency between input studies and reviews, and to provide additional information for interpreting the findings reported in each document.

\subsubsection{Adding PICO tags}
\label{sec:pico_tagging}
The Populations, Interventions, and Outcomes of interest are a common way of representing clinical knowledge \citep{Huang2006EvaluationOP}. Recent work \citep{nye2020understanding} has found that the Comparator is rarely mentioned explicitly, so we exclude it from our dataset.
Previous summarization work has shown that tagging salient entities, especially PIO elements \citep{Wallace2020GeneratingN}, can improve summarization performance \citep{ nallapati-etal-2016-abstractive,Nallapati2016AbstractiveTS}, so we mark PIO elements with special tokens added to our model vocabulary: \texttt{<pop>}, \texttt{</pop>}, \texttt{<int>}, \texttt{</int>}, \texttt{<out>}, and \texttt{</out>}.

Using the EBM-NLP corpus \citep{ebmnlp}, a crowd-sourced collection of PIO tags,\footnote{EBM-NLP contains high quality, crowdsourced and expert-tagged PIO spans in clinical trial abstracts. See App.~\ref{sec:overview} for a comparison to other PICO datasets.
} we train a token classification model \citep{huggingface} to identify these spans in our study and review documents. These span sets are denoted $\boldsymbol{P} = \{P_1, P_2,..., P_{\bar{P}}\}$, $\boldsymbol{I} = \{I_1, I_2,..., I_{\bar{I}}\}$ and $\boldsymbol{O} =  \{O_1, O_2,..., O_{\bar{O}}\}$. At the level of each review, we perform a simple aggregation over these elements. Any P, I, or O span fully contained within any other span of the same type is removed from these sets (though they remain tagged in the text). Removing these contained elements reduces the number of duplicates in our structured representation. Our dataset has an average of 3.0 P, 3.5 I, and 5.4 O spans per review.

\subsubsection{Adding Evidence Inference}
\label{sec:evidence_inference}

We predict the direction of evidence associated with every Intervention-Outcome (I/O) pair found in the review abstract.
Taking the product of each $I_i$ and $O_j$ in the sets $\boldsymbol{I}$ and $\boldsymbol{O}$ yields all possible I/O pairs, and each I/O pair is associated with an evidence direction $d_{ij}$, which can take on one of the values in $\{$\increases, \nochange, \decreases$\}$. For each I/O pair, we also derive a sentence $s_{ij}$ from the document supporting the $d_{ij}$ classification. 
Each review can therefore be represented as a set of tuples $\boldsymbol{T}$ of the form $(I_i, O_j, s_{ij}, d_{ij})$ and cardinality $\bar{I} \times \bar{O}$. See Tab.~\ref{tab:sample_pico} for examples. For modeling, as in PICO tagging, we surround supporting sentences with special tokens \texttt{<evidence>} and \texttt{</evidence>}; and append the direction class with a \texttt{<sep>} token.

We adapt the Evidence Inference (EI) dataset and models \citep{evidenceinference} for labeling. The EI dataset is a collection of RCTs, tagged PICO elements, evidence sentences, and overall evidence direction labels \increases, \nochange, or \decreases. 
The EI models are composed of 1) an evidence identification module which identifies an evidence sentence, and 2) an evidence classification module for classifying the direction of effectiveness. The former is a binary classifier on top of SciBERT, whereas the latter is a softmax distribution over effectiveness directions.
Using the same parameters as \citet{evidenceinference}, we modify these two modules to function solely over I and O spans.\footnote{\citet{nye2020understanding} found that removing Comparator elements improved classification performance from 78.0 F1 to 81.4 F1 with no additional changes or hyper-parameter tuning.} The resulting 354k EI classifications for our reviews are 13.4\% \decreases, 57.0\% \nochange, and 29.6\% \increases. Of the 907k classifications over input studies, 15.7\% are \decreases, 60.7\% \nochange, and 23.6\% \increases. Only 53.8\% of study classifications match review classifications, highlighting the prevalence and challenges of contradictory data.

\subsection{Clustering and train / test split}
\label{subsection:clustering_def}
Reviews addressing overlapping research questions or providing updates to previous reviews may share input studies and results in common, e.g., a review studying the effect of Vitamin B12 supplementation on B12 levels in older adults and a review studying the effect of B12 supplementation on heart disease risk will cite similar studies.
To avoid the phenomenon of learning from test data, we cluster reviews before splitting into train, validation, and test sets. We compute SPECTER paper embeddings \cite{specter} using the title and abstract of each review, and perform agglomerative hierarchical clustering using the scikit-learn library \citep{sklearn_api}. This results in 200 clusters, which we randomly partition into 80/10/10 train/development/test sets.

\subsection{Dataset statistics}

The final dataset consists of \numreviews reviews and \numstudies studies. Each review in the dataset summarizes an average of 23 studies, ranging between 1--401 
studies. See Tab.~\ref{tab:stats} for statistics, and Tab.~\ref{tab:comparable_dataset_sizes} for a comparison to other datasets. The median review has 6.7K input tokens from its input studies, while the average has 9.4K tokens (a few reviews have lots of studies). We restrict the input size when modeling to 25 studies, which reduces the average input to 6.6K tokens without altering the median.

\begin{table}[tb!]
    \centering
    \small
    \begin{tabular}{ll}
        \toprule
         Dataset statistics & \litreview \\
        \midrule
         Total reviews & \numreviews \\
         Total input studies & \numstudies \\
         Median number of studies per review & 17 \\
         Median number of reviews per study & 1 \\
         Average number of reviews per study & 1.9 \\
         Median \background sentences per review & 3 \\
         Median \target sentences per review & 2\\
        \bottomrule
    \end{tabular}
    \caption{\litreview dataset statistics.
    }
    \label{tab:stats}
\end{table}

\begin{table}[tb!]
    \centering
    \small
    \begin{tabular}{llL{15mm}L{15mm}}
        \toprule
         Dataset    & Docs      & Tokens per Summary & Tokens per Document \\
        \midrule
         DUC '03/'04   & 320       & 109.6 & 4636.2 \\
         TAC 2011   & 176       & 99.7 & 4695.7 \\
         WikiSum & 2,332,000 & $10^1$--$10^3$ & $10^2$--$10^6$ \\
         Multi-News & 56,216    & 263.7 & 2103.4 \\
         \litreview & 470,402   & 61.3  & 365.3 \\
        \bottomrule
    \end{tabular}
    \caption{A comparison of \MDS datasets; adapted from \citet{multinews}. Datasets are DUC '03/'04$^{\ref{footnote:duc}}$, TAC 2011 \citep{tac2011}, WikiSum \citep{2018generating}, and Multi-News \cite{multinews}.     Note: WikiSum only provides ranges, not exact size. 
    }
    \vspace{-1.5em}
    \label{tab:comparable_dataset_sizes}
\end{table}

Fig.~\ref{fig:delay} shows the temporal distribution of reviews and input studies in \litreview. We observe that though reviews in our dataset have a median publication year of 2016, the studies cited by these reviews are largely from before 2010, with a median of 2007 and peak in 2009. This citation delay has been observed in prior work \citep{Shojania2007HowQD, Beller2013AreSR}, and further illustrates the need for automated or assisted reviews. 

%% file: sections/models.tex
\section{Experiments}
\label{sec:models}

\begin{figure}[t!]
    \centering
    \includegraphics[width=\linewidth]{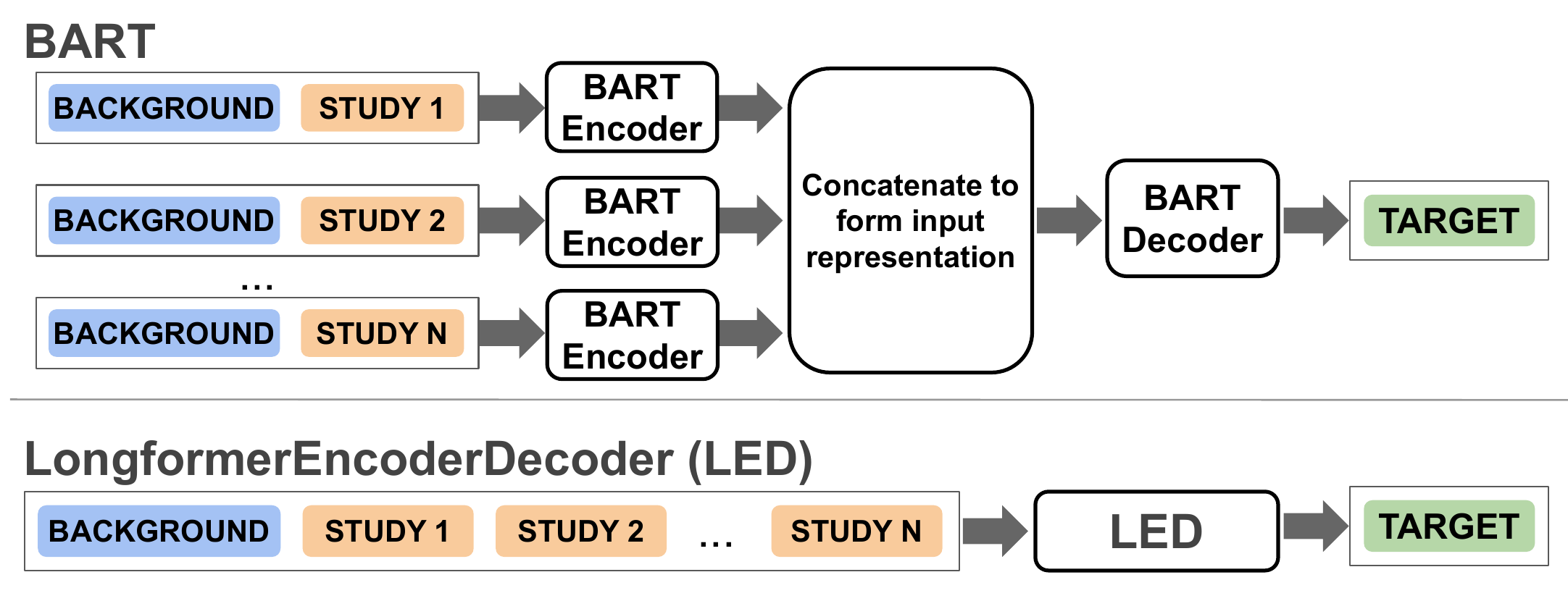}
    \caption{Two input encoding configurations.
    Above: LongformerEncoderDecoder (\longbart), where all input studies are appended to the \background and encoded together. 
    Below: In the \standard configuration, each input study is encoded independently with the review \background. These are concatenated to form the input encoding. 
    }
    \label{fig:model_sketch}
    \vspace{-1.5em}
\end{figure}

We experiment with a \texttotext task formulation (Fig.~\ref{fig:tasks}). 
The model input consists of the \background statement and study abstracts; the output is the \target statement. We also investigate the use of the 
structured form described in \S\ref{sec:evidence_inference} for a supplementary \tabtotab task, where given inputs of I/O pairs from the review; the model tries to predict the evidence direction. We provide initial results for the \tabtotab task, although we consider this an area in need of active research.

\subsection{Texts-to-text task}
\label{sec:text2text}
Our approach leverages BART \citep{BART}, a \textit{seq2seq} autoencoder. 
Using BART, we encode the \background and input studies as in Fig.~\ref{fig:model_sketch}, and pass these representations to a decoder. 
Training follows a standard auto-regressive paradigm used for building summarization models. 
In addition to PICO tags 
(\S\ref{sec:pico_tagging}), we augment the inputs by surrounding the background and each input study with special tokens \texttt{<background>}, \texttt{</background>}, and \texttt{<study>}, \texttt{</study>}.

For representing multiple inputs, we experiment with two configurations: 
one leveraging BART with independent encodings of each input, and LongformerEncoderDecoder (\longbart)~\citep{longformer} which can encode  long inputs of up to 16K tokens.  
For the \standard configuration, each study abstract is appended to the \background statement and encoded independently. 
These representations are concatenated together to form the input to the decoder layer. 
In the \standard configuration, interactions happen only in the decoder.
For the \longbart configuration, the input sequence starts with the \background statement followed by a concatenation of all input study abstracts.
The \background representation is shared among all input studies; global attention allows interactions between studies, and a sliding attention window of 512 tokens allows each token to attend to its neighbors.

We train a \standard-base model, with hyperparameters described in App.~\ref{sec:hyperparameters}.
We report experimental results in Tab.~\ref{tab:text_to_text_results}. In addition to ROUGE \citep{Lin2004ROUGEAP}, we also report two metrics derived from evidence inference: \jsdmetric and F1. We describe the intuition and computation of the \jsdmetric metric in Section \ref{sec:jsd}; because it is a distance metric, lower \jsdmetric is better. For F1, we use the EI classification module to identify evidence directions for both the generated and target summaries. Using these classifications, we report a macro-averaged F1 over the class agreement between the generated and target summaries \cite{sklearn_api}. For example generations, see Tab.~\ref{tab:examples} in App.~\ref{appendix:examples}.

\subsection{Table-to-table task}
\label{sec:table2table}
An end user of a review summarization system may be interested in specific results from input studies (including whether they agree or contradict) rather than the high level conclusions available in \target statements. 
Therefore, we further experiment with structured input and output representations that attempt to capture results from individual studies. As described in \S\ref{sec:evidence_inference}, the structured representation of each review or study is a tuple
of the form $(I_i, O_j, s_{ij}, d_{ij})$. It is important to note that
we use the same set of $\boldsymbol{I}$s and $\boldsymbol{O}$s
from the review to predict evidence direction from all input studies. 

Borrowing from the ideas of \citep{Raffel2020ExploringTL}, we formulate our classification task as a text generation task, and train the models described in Section \ref{sec:text2text} to generate one of the classes in $\{$\increases, \nochange, \decreases$\}$. Using the EI classifications from \ref{sec:evidence_inference}, we compute an F-score macro-averaged over the effect classes (Tab.~\ref{tab:table_to_table_results}). 
We retain all hyperparameter settings other than reducing the maximum generation length to 10.


We stress that this is a preliminary effort to demonstrate feasibility rather than completeness \hspace*{.05em} --- our results in Tab.~\ref{tab:table_to_table_results} are promising but the underlying technologies for building the structured data: PICO tagging, co-reference resolution, and PICO relation extraction, are currently weak \citep{nye2020understanding}. Resorting to using the full cross-product of Interventions and Outcomes results in duplicated I/O pairs as well as potentially spurious pairs that do not correspond to actual I/O pairs in the review. 

\begin{table}[t!]
    \centering
    \small
    \begin{tabular}{llllll}
    \toprule
    Model         & R-1     & R-2   & R-L   & $\Delta$EI$\downarrow$ & F1 \\
    \midrule
    \standard     & 27.56   & 9.40  & 20.80 & .459 & 46.51 \\
    \longbart     & 26.89   & 8.91  & 20.32 & .449 & 45.00 \\
    \bottomrule
    \end{tabular}
    \caption{Results for the \texttotext setting. 
    We report ROUGE, $\Delta$EI (\S~\ref{sec:jsd}), and macro-averaged F1-scores. 
    }
    \label{tab:text_to_text_results}
\end{table}

\begin{table}[t!]
    \centering
    \small
    \begin{tabular}{llll}
    \toprule
    Model       & P     & R     & F1 \\
    \midrule
    \standard   & 50.31 & 67.98 & 65.89 \\
    \bottomrule
    \end{tabular}
    \caption{Results for the \tabtotab setting. We report macro-averaged precision, recall, and F-scores.
    }
    \label{tab:table_to_table_results}
\end{table}

\subsection{\jsdmetric metric}
\label{sec:jsd}

Recent work in summarization evaluation has highlighted the weaknesses of ROUGE for capturing factuality of generated summaries, and has focused on developing automated metrics more closely correlated with human-assessed factuality and quality \citep{bert-score,qags,falke-etal-2019-ranking}. In this vein, we modify a recently proposed metric based on EI classification distributions \citep{Wallace2020GeneratingN}, intending to capture the agreement of Is, Os, and EI directions between input studies and the generated summary.  %

For each I/O tuple $(I_i, O_j)$, the predicted direction $d_{ij}$ is actually a distribution of probabilities over the three direction classes $P_{ij} = (p_{increases}, p_{decreases}, p_{no\_change})$. If we consider this distribution for the gold summary ($P_{ij}$) and the generated summary ($Q_{ij}$), we can compute the Jensen-Shannon Distance (JSD) \cite{Lin1991DivergenceMB}, a bounded score between $[0, 1]$, between these distributions. For each review, we can then compute a summary JSD metric, which we call \jsdmetric, as an average over the JSD of each I/O tuple in that review:
\begin{equation}
    \sum_{i=1}^{\bar{I}}\sum_{j=1}^{\bar{J}} JSD(P_{ij},Q_{ij})
\end{equation}
\noindent Different from \citet{Wallace2020GeneratingN}, \jsdmetric is an average over all outputs, attempting to capture an overall picture of system performance,\footnote{\citet{Wallace2020GeneratingN} only report correlation of a related metric with human judgments.
} and our metric retains the directionality of \increases and \decreases, as opposed to collapsing them together.

To facilitate interpretation of the \jsdmetric metric, we offer a degenerate example. Given the case where all direction classifications are certain, and the probability distributions $P_{ij}$ and $Q_{ij}$ exist in the space of $(1,0,0)$, $(0,1,0)$, or $(0,0,1)$, \jsdmetric takes on the following values at various levels of consistency between $P_{ij}$ and $Q_{ij}$ for the input studies: 
\begin{equation*}
\begin{array}{rl}
    \textrm{100\% consistent} & $\jsdmetric = 0.0$  \\
    \textrm{50\% consistent} & $\jsdmetric = 0.42$ \\
    \textrm{0\% consistent} & $\jsdmetric = 0.83$
\end{array}
\end{equation*}

\noindent In other words, in both the standard BART and \longbart setting, the evidence directions predicted in relation to the generated summary are slightly less than 50\% consistent with the direction predictions produced relative to the gold summary.

\subsection{Human evaluation \& error analysis}
\label{sec:humaneval}

We randomly sample 150 reviews from the test set for manual evaluation. For each generated and gold summary, we annotate the primary effectiveness direction in the summary to the following classes: (i) \textit{increases}: intervention has a positive effect on the outcome; (ii) \textit{no\_change}: no effect, or no difference between the intervention and the comparator; (iii) \textit{decreases}: intervention has a negative effect on the outcome; (iv) \textit{insufficient}: insufficient evidence is available; (v) \textit{skip}: the summary is disfluent, off topic, or does not contain information on efficacy.

Here, \textit{increases}, \textit{no\_change}, and \textit{decreases} correspond to the EI classes, while we introduce \textit{insufficient} to describe cases where insufficient evidence is available on efficacy, and \textit{skip} to describe data or generation failures. 
Two annotators provide labels, and agreement is computed over 50 reviews (agreement: 86\%, Cohen's $\kappa$: 0.76). Of these, 17 gold summaries lack an efficacy statement, and are excluded from analysis. 
Tab.~\ref{tab:human_eval} shows the confusion matrix for the sample. Around 50\% (67/133) of generated summaries have the same evidence direction as the gold summary. Most confusions happen between \textit{increases}, \textit{no\_change}, and \textit{insufficient}. 

\begin{table}[t!]
    \centering
    \small
    \begin{tabular}{r|cccc}
    \toprule
     & \multicolumn{4}{c}{Gold} \\
    Generated & \textit{inc.} & \textit{no\_c.} & \textit{dec.} & \textit{insuff.} \\
    \midrule
    \textit{increases} & 56 & 3 & 1 & 19 \\
    \textit{no\_change} & 13 & 1 & 1 & 10 \\
    \textit{decreases} & 0 & 0 & 5 & 2 \\
    \textit{insufficient} & 5 & 1 & 0 & 5\\
    \textit{skip} & 8 & 0 & 0 & 3 \\
    \bottomrule
    \end{tabular}
    \caption{Confusion matrix for human evaluation results}
    \label{tab:human_eval}
    \vspace{-1em}
\end{table}

Tab.~\ref{tab:contradictions} shows how individual studies can provide contradictory information, some supporting a positive effect for an intervention and some observing no or negative effects. EI may be able to capture some of the differences between these input studies. From observations on limited data: while studies with positive effect tend to have more EI predictions that were \increases or \decreases, those with no or negative effect tended to have predictions that were mostly \nochange. However, more work is needed to better understand how to capture these directional relations and how to aggregate them into a coherent summary. 

\begin{table}[t!]
    \small
    \centering
    \begin{tabular}{L{17mm}L{51mm}}
        \toprule
        Effectiveness & Example statements from studies \\
        \midrule
        Positive effect & Adjuvant vinorelbine plus cisplatin \textbf{extends survival} in patients with completely resected NSCLC... \\ \cmidrule{2-2}
            & Our results suggest that patients with NSCLC at pathologic stage I who have undergone radical surgery \textbf{benefit} from adjuvant chemotherapy. \\
        \midrule
        No effect \textit{or} negative effect & \textbf{No survival benefit} for CAP vs no-treatment control was found in this study. Therefore, adjuvant therapy with CAP \textbf{should not be recommended} for patients with resected early-stage non-small cell lung cancer . \\ \cmidrule{2-2}
            & On the basis of this trial , adjuvant therapy with CAP \textbf{should not be recommended} for patients with resected stage I lung cancer . \\
        \bottomrule
    \end{tabular}
    \caption{Text from the input studies to \citet{Petrelli2013NoncancerrelatedMA}, a review investigating the effectiveness of cisplatin-based (CAP) chemotherapy for non-small cell lung cancer (NSCLC). Input studies vary in their results, with some stating a positive effect for adjuvant chemotherapy, and some stating no survival benefit.}
    \label{tab:contradictions}
    \vspace{-1em}
\end{table}

%% file: sections/related_work.tex
\section{Related Work}

NLP for scientific text has been gaining interest recently
with work spanning the whole NLP pipeline: datasets (S2ORC \citep{s2orc}, CORD-19 \citep{cord19}), pretrained transformer models (SciBERT \citep{beltagy-etal-2019-scibert}, BioBERT \citep{Lee2020BioBERTAP}, ClinicalBERT \citep{Huang2019ClinicalBERTMC}, SPECTER \citep{specter}), NLP tasks like NER \citep{ebmnlp, Li2016BioCreativeVC}, relation extraction \citep{scirex, scierc, chemprot}, 
QA \citep{mediqa}, NLI \citep{mednli, scitail}, summarization \citep{scitldr, scisumm}, claim verification \cite{scifact}, and more. \litreview adds a \MDS dataset to the scientific document NLP literature. 

A small number of \MDS datasets are available for other domains, including MultiNews \citep{multinews}, WikiSum \citep{2018generating}, 
and Wikipedia Current Events \citep{wcep}.
Most similar to \litreview is MultiNews, where multiple news articles about the same event are summarized into one short paragraph. 
Aside from being in a different textual domain (scientific vs. newswire), one unique characteristic of \litreview compared to existing datasets is that \litreview input documents have contradicting evidence. Modeling in other domains has typically focused on straightforward applications of single-document summarization to the multi-document setting \citep{lebanoff-etal-2018-adapting,Zhang2018TowardsAN}, although some methods explicitly model multi-document structure using semantic graph approaches \citep{Baumel2018QueryFA,liu-lapata-2019-hierarchical,li-etal-2020-leveraging}.

In the systematic review domain, work has typically focused on information retrieval \citep{boudin-etal-2010-clinical,Ho2016DevelopmentOA,Znaidi2015AnsweringPC,Schoot2020ASReviewOS}, extracting findings \citep{lehman-etal-2019-inferring,evidenceinference,nye2020understanding}, and quality assessment \citep{Marshall2015AutomatingRO,Marshall2016RobotReviewerEO}. Only recently in \citet{Wallace2020GeneratingN} and this work has consideration been made for approaching the entire system as a whole. We refer the reader to App. \ref{sec:overview} for more context regarding the systematic review process.

%% file: sections/discussion.tex
\section{Discussion}

Though \MDS has been explored in the general domain, biomedical text poses unique challenges such as the need for domain-specific vocabulary and background knowledge. 
To support development of biomedical \MDS systems, we release the \litreview dataset. \litreview contains summaries and documents derived from biomedical literature, and can be used to study literature review automation, a pressing real-world application of \MDS.

We define a \textit{seq2seq} modeling task over this dataset, as well as a structured task that incorporates prior work on modeling biomedical text \cite{ebmnlp, evidenceinference}. We show that although generated summaries tend to be fluent and on-topic, they only agree with the evidence direction in gold summaries around half the time, leaving plenty of room for improvement. This observation holds both through our \jsdmetric metric and through human evaluation of a small sample of generated summaries. 
Given that only 54\% of study evidence directions agree with the evidence directions of their review, modeling contradiction in source documents may be key to improving upon existing summarization methods.

\paragraph{Limitations} Challenges in co-reference resolution and PICO extraction limit our ability to generate accurate PICO labels at the document level. Errors compound at each stage: PICO tagging, taking the product of Is and Os at the document level, and predicting EI direction. 
Pipeline improvements are needed to bolster overall system performance and increase our ability to automatically assess performance via automated metrics like \jsdmetric. 
Relatedly, automated metrics for summarization evaluation can be difficult to interpret, as the intuition for each metric must be built up through experience. 
Though we attempt to facilitate understanding of \jsdmetric by offering a degenerate example, more exploration is needed to understand how a practically useful system would perform on such a metric.

\paragraph{Future work} Though we demonstrate that \textit{seq2seq} approaches are capable of producing fluent and on-topic review summaries, there are significant opportunities for improvement. 
Data improvements include improving the quality of summary targets and intermediate structured representations (PICO tags and EI direction). Another opportunity lies in linking to structured data in external sources such as various clinical trial databases\footnote{\href{https://clinicaltrials.gov/}{https://clinicaltrials.gov/}}$^,$\footnote{\href{https://www.clinicaltrialsregister.eu/}{https://www.clinicaltrialsregister.eu/}}$^,$\footnote{\href{https://www.gsk-studyregister.com/}{https://www.gsk-studyregister.com/}} rather than relying solely on PICO tagging. 
For modeling, we are interested in pursuing joint retrieval and summarization approaches \citep{Lewis2020PretrainingVP}.
We also hope to explicitly model the types of contradictions observed in Tab.~\ref{tab:contradictions}, such that generated summaries can capture nuanced claims made by individual studies.

%% file: sections/conclusions.tex
\section{Conclusion}

Given increasing rates of publication, multi-document summarization, or the creation of literature reviews, has emerged as an important NLP task in science. The urgency for automation technologies has been magnified by the COVID-19 pandemic, which has led to both an accelerated speed of publication \citep{Horbach2020PandemicPM} as well as proliferation of non-peer-reviewed preprints which may be of lower quality \citep{Lachapelle2020COVID19PA}. 
By releasing \litreview, we provide a \MDS dataset that can help to address these challenges. Though we demonstrate that our \MDS models can produce fluent text, our results show that there are significant outstanding challenges that remain unsolved, such as PICO tuple extraction, co-reference resolution, and evaluation of summary quality and faithfulness in the multi-document setting. We encourage others to use this dataset to better understand the challenges specific to \MDS in the domain of biomedical text, and to push the boundaries on the real world task of systematic review automation.

%% file: sections/broader_impact.tex
\section*{Ethical Concerns and Broader Impact}
\label{sec:ethics_broader_impact}

We believe that automation in systematic reviews has great potential value to the medical and scientific community;
our aim in releasing our dataset and models is to facilitate research in this area.
Given unresolved issues in evaluating the factuality of summarization systems, as well as a lack of strong guarantees about what the summary outputs contain, we do not believe that such a system is ready to be deployed in practice. Deploying such a system now would be premature, as without these guarantees we would be likely to generate plausible-looking but factually incorrect summaries, an unacceptable outcome in such a high impact domain. We hope to foster development of useful systems with correctness guarantees and evaluations to support them.

%% file: sections/appendix.tex
\appendix

\begin{figure}[t!]
    \centering
    \includegraphics[width=\linewidth]{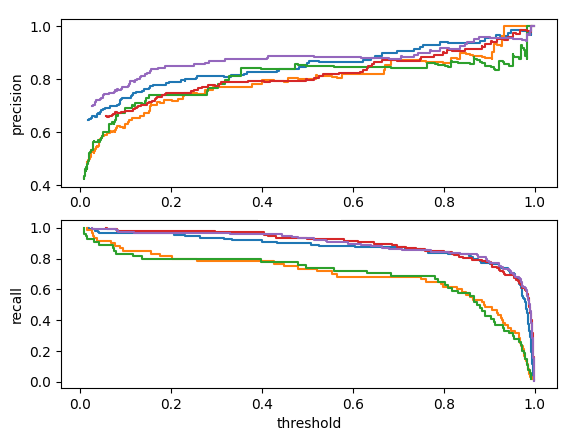}
    \caption{Five fold cross-validation results from training a binary SciBERT classifier on the annotations. Precisions increase following a logistic curve over threshold choices; recalls decrease.}
    \label{fig:dataset:abstract_threshold}
\end{figure}

\begin{table}[t!]
    \centering
    \small
    \begin{tabular}{lccc}
    \toprule
       Sentence class & P & R & F\\
    \midrule
BACKGROUND & 0.972  & 0.952 & 0.958\\
RECOMMENDATION & 0.418  & 0.296 & 0.338\\
EVIDENCE\_QUALITY & 0.580 & 0.528 & 0.550 \\
EFFECT & 0.752 & 0.800 & 0.774\\
METHODS & 0.938 & 0.944  & 0.940\\
GOAL & 0.916 & 0.936 & 0.924\\
DETAILED\_FINDINGS & 0.856  & 0.860 & 0.858\\
ETC & 0.406 & 0.322 & 0.338 \\
FURTHER\_STUDY & 0.756 & 0.864 & 0.804\\
    \bottomrule
    \end{tabular}
    \caption{Precision, Recall, and F1-scores for all annotation classes, averaged over five folds of cross validation.}
    \label{tab:detailed_sentence_breakdown_scores}
\end{table}

\begin{table*}[t!]
    \centering
    \scriptsize
    \begin{tabular}{l|p{10mm}p{6mm}p{10mm}p{12mm}p{12mm}p{15mm}p{12mm}p{8mm}p{6mm}}
    \toprule
 & BACK-GROUND & GOAL & METHODS & DETAILED\_FINDINGS & FURTHER\_STUDY & RECOMMEND-ATION & EVIDENCE\_QUALITY & EFFECT & ETC\\
 \midrule
BACKGROUND &  0.952 &  0.017 &  0.008 &  0.003 &  0.0 &  0.003 &  0.003 &  0.008 &  0.006\\
GOAL &  0.009 &  0.935 &  0.056 &  0.0 &  0.0 &  0.0 &  0.0 &  0.0 &  0.0\\
METHODS &  0.003 &  0.011 &  0.944 &  0.034 &  0.0 &  0.0 &  0.0 &  0.006 &  0.001\\
DETAILED\_FINDINGS &  0.001 &  0.0 &  0.027 &  0.862 &  0.0 &  0.001 &  0.027 &  0.079 &  0.002\\
FURTHER\_STUDY &  0.0 &  0.0 &  0.0 &  0.0 &  0.857 &  0.061 &  0.01 &  0.02 &  0.051\\
RECOMMENDATION &  0.0 &  0.0 &  0.021 &  0.021 &  0.277 &  0.298 &  0.021 &  0.277 &  0.085\\
EVIDENCE\_QUALITY &  0.0 &  0.0 &  0.016 &  0.227 &  0.008 &  0.016 &  0.523 &  0.148 &  0.062\\
EFFECT &  0.004 &  0.0 &  0.002 &  0.114 &  0.013 &  0.015 &  0.032 &  0.8 &  0.021\\
ETC &  0.051 &  0.051 &  0.064 &  0.128 &  0.077 &  0.038 &  0.09 &  0.218 &  0.282\\
\bottomrule
\end{tabular}
    \caption{Full 9-class sentence classification confusion matrix, averaged over five folds of cross validation.}
    \label{tab:senteces_confusion_matrix}
\end{table*}

\begin{table}[t!]
    \centering
    \scriptsize
    \begin{tabular}{p{15mm}p{54mm}}
        \toprule
        \small{Label} & \small{Sentence} \\
        \midrule
        \background & IMPORTANCE OF THE FIELD Cobalamin ( vitamin B12 ) deficiency is particularly common in the {\pop{elderly ( $>$ 15 \% ) }} . \\
        \hline
        \background & Management of cobalamin deficiency with {\inte{cobalamin injections}} is well codified at present , but new routes of {\inte{cobalamin}} administration ( oral and nasal ) are being studied , especially oral {\inte{cobalamin}} therapy for food-cobalamin malabsorption . \\
        \hline
        \background & AREAS COVERED IN THIS REVIEW The objective of this review is to evaluate the {\out{efficacy of}} {\inte{oral cobalamin treatment}} in {\pop{elderly patients .}} \\
        \hline
        \other & To reach this objective , PubMed data were systematic ally search ed for {\pop{English and French articles published from January 1990 to July 2008 .}} \\
        \hline
        \other & Data from our research group on cobalamin deficiency ( Groupe d'Etude des CAREnce vitamine B12 - CARE B12 ) were also analyzed . \\
        \hline
        \other & WHAT THE READER WILL GAIN Three prospect i ve r and omized studies , a systematic review by the Cochrane group and five prospect i ve cohort studies were found and provide evidence that {\inte{oral cobalamin treatment}} may adequately treat cobalamin deficiency . \\
        \hline
        \target & The {\out{efficacy}} was particularly highlighted when looking at the marked improvement in {\out{serum vitamin B12 levels and hematological parameters}} , for example {\out{hemoglobin level , mean erythrocyte cell volume and reticulocyte count .}} \\
        \hline
        \other & The {\out{effect}} of {\inte{oral cobalamin treatment}} in {\pop{patients}} presenting with {\out{severe neurological manifestations}} has not yet been adequately documented . \\
        \hline
        \target & \phantom{}{\inte{Oral cobalamin treatment}} avoids the {\out{discomfort , inconvenience and cost}} of monthly injections . \\
        \hline
        \target & TAKE HOME MESSAGE Our experience and the present analysis support the use of {\inte{oral cobalamin therapy}} in clinical practice \\
       
        \bottomrule
    \end{tabular}
    \caption{Example review abstract from \citet{Andrs2010EfficacyOO} with predicted sentence labels. Spans corresponding to {\pop{Population}}, {\inte{Intervention}}, and {\out{Outcome}} elements are tagged and surrounded with special tokens.}
    \label{tab:sent_example_full}
\end{table}

\section{MeSH Filtering}
\label{appendix:mesh_filtering}

For each candidate review, we extract its cited papers and identify the study type of each cited paper using MeSH publication type,\footnote{\href{https://www.nlm.nih.gov/mesh/pubtypes.html}{https://www.nlm.nih.gov/mesh/pubtypes.html}} keeping only studies that are clinical trials, cohort studies, and/or observational studies (see Appendix \ref{appendix:suitbility_mesh_terms} for full list of MeSH terms). We exclude case reports, which usually report findings on one or a small number of individuals.
We observe that publication type MeSH terms tend to be under-tagged.\footnote{From a cursory inspection of a random sample of studies, this problem seems to be widespread.} Therefore, we also use ArrowSmith trial labels \cite{Cohen2015AutomatedCR,Shao2015AggregatorAM} and a keyword heuristic (the span ``randomized'' occurring in the title or abstract) to identify additional RCT-like studies.\footnote{RCTs provide the highest quality of evidence so we strive to include as many as possible as inputs in our dataset.} 
Candidate reviews are culled to retain only those that cite at least one suitable study and no case studies, leaving us with 30K reviews.

\subsection{Suitability MeSH Terms}
\label{appendix:suitbility_mesh_terms}

We use the following publication type MeSH terms to decide whether a review's input document is a study of interest:
\begin{enumerate}[noitemsep]
    \item `Clinical Study'
    \item `Clinical Trial'
    \item `Controlled Clinical Trial'
    \item `Randomized Controlled Trial'
    \item `Pragmatic Clinical Trial'
    \item `Clinical Trial, Phase I'
    \item `Clinical Trial, Phase II'
    \item `Clinical Trial, Phase III'
    \item `Clinical Trial, Phase IV'
    \item `Equivalence Trial'
    \item `Comparative Study'
    \item `Observational Study'
    \item `Adaptive Clinical Trial'
\end{enumerate}

And we exclude any reviews citing studies with the following publication type MeSH terms:
\begin{enumerate}[noitemsep]
    \item `Randomized Controlled Trial, Veterinary'
    \item `Clinical Trial, Veterinary'
    \item `Observational Study, Veterinary'
    \item `Case Report'
\end{enumerate}


\section{Suitability Annotation}
\label{appendix:suitability_annotation}
The annotation guidelines for review suitability are given below. Each annotator was tasked with an initial round of annotation, followed by a round of review, then further annotation.

\subsection{Suitability Guidelines}
A systematic review is a document resulting from an in-depth search and analysis of all the literature relevant to a particular topic. We are interested in systematic reviews of medical literature, specifically those that assess varying treatments and the outcomes associated with them.

There are many different types of reviews, and many types of documents that look like reviews. We need to identify only the “correct” types of reviews. Sometimes this can be done from the title alone, sometimes one has to read the review itself.

The reviews we are interested in:
\begin{itemize}
    \item Must study a human population (no animal, veterinary, or environmental studies)
    \item Must review studies involving multiple participants. We are interested in reviews of trials or cohort studies. We are *not* interested in reviews of case studies - which describe one or a few specific people. 
    \item Must study an explicit population or problem (P from PICO)
    \begin{itemize}
        \item Example populations: women $>$ 55 old with breast cancer, migrant workers, elementary school children in Spokane, WA, etc.
    \end{itemize}
    \item Must compare one or more medical interventions
    \begin{itemize}
        \item Example interventions: drugs, vaccines, yoga, therapy, surgery, education, annoying mobile device reminders, professional naggers, personal trainers, and more! Note: placebo / no intervention is a type of intervention.
        \item Comparing the effectiveness of an intervention against no intervention is okay.
        \item Combinations of interventions count as comparisons (e.g. yoga vs. yoga + therapy). 
        \item Two different dosages also count (e.g. 500ppm fluoride vs 1000ppm fluoride in toothpaste).
    \end{itemize}
    \begin{itemize}
        \item Must have an explicit outcome measure
        \item Example outcome measures: survival time, frequency of headaches, relief of depression, survey results, and many other possibilities.
    \end{itemize}
    \item The outcome measure must measure the effectiveness of the intervention.
\end{itemize}

\section{Suitability Classifier}
\label{appendix:suitability_classifier}

Four annotators with biomedical background labeled 879 reviews sampled from the candidate pool (572 suitable, 307 not, Cohen's Kappa: 0.55) according to the suitability criteria (guidelines in Appendix \ref{appendix:suitability_annotation}). We aim to include reviews that perform an aggregation over existing results, such as reporting on how a medical or social intervention affects a group of people, while excluding reviews that make new observations, such as identifying novel disease co-morbidities or those that synthesize case studies.

For our suitability classifier, we finetune SciBERT~\citep{beltagy-etal-2019-scibert} using standard parameters; using five-fold cross validation we find that a threshold of 0.75 provides a precision of greater than 80\% while maintaining an adequate recall (Figure \ref{fig:dataset:abstract_threshold}). 

Though there are a fairly large number of false positives by this criteria, we note that these false positive documents \textit{are} generally reviews; however, they may not investigate an intervention, or may not have suitable target statements.
In the latter case, target identification described in \S~\ref{subsection:sentence_labeling_def} helps us further refine and remove these false positives from the final dataset.

\section{Sentence Annotation}
\label{appendix:evidence_sentence_annotation}

Sentence annotation guidelines and detailed scores are below. Each annotator was tasked with annotating 50-100 sentences, followed by a round of review, before being asked to annotate more.

\subsection{Sentence Annotation Guidelines}
A systematic review is a document resulting from an in-depth search and analysis of all the literature relevant to a particular topic. We are interested in systematic reviews of medical literature, specifically those that assess varying treatments and the outcomes associated with them. Ignore any existing labels; these are automatically produced and error prone. If something clearly fits into more than one category, separate the labels by commas (annoying, we know, but it can be important). For sentences that are incorrectly broken in a way that makes them difficult to label, skip them (you can fix them, but they’ll be programmatically ignored). For reviews that don’t meet suitability guidelines, also skip them. We want to identify sentences within these reviews as belonging to one of several categories:

\begin{itemize}
    \item \textbf{BACKGROUND}: Any background information not including goals.
    \item \textbf{GOAL}: A high level goal sentence, describing the aims or purposes of the review.
    \item \textbf{METHODS}: Anything describing the particular strategies or techniques for conducting the review. This includes methods for finding and assessing appropriate studies to include, e.g., the databases searched or other characteristics of the searched literature. A characteristic might be a study type, it might be other details, such as criteria involving the study participants, what interventions (treatments) were studied or compared, or what outcomes are measured in those studies. This may also include whether or not a meta-analysis is performed.
    \item \textbf{DETAILED\_FINDINGS}: Any sections reporting study results, often includes numbers, p-values, etc. These will frequently include statements about a subset of the trials or the populations.
    \item GENERAL FINDINGS: There are four types of general findings we would like you to label. These do not include things like number of patients, or a p-value (that’s DETAILED FINDINGS). Not all of these four subtypes will always be present in a paper’s abstract. Some sentences will contain information about more than one subtype, and some sentences can contain information about some of these subtypes as well as DETAILED FINDINGS. 
    \begin{itemize}
        \item \textbf{EFFECT}: Effect of the intervention, may include a statement about significance. These can cover a wide range of topics, including public health or policy changes.
        \item \textbf{EVIDENCE\_QUALITY}: Commentary about the strength or quality of evidence pertaining to the intervention. 
        \item \textbf{FURTHER\_STUDY}: These statements might call for more research in a particular area, and can include hedging statements, e.g.:
        \begin{itemize}
            \item “More rigorously designed longitudinal studies with standardized definitions of periodontal disease and vitamin D are necessary.”
            \item “More research with larger sample size and high quality in different nursing educational contexts are required.”
            \item “However, this finding largely relies on data from observational studies; high-quality RCTs are warranted because of the potential for subject selection bias.”
        \end{itemize} 
        \item \textbf{RECOMMENDATION}: Any kind of clinical or policy recommendation, or recommendations for use in practice. This must contain an explicit recommendation, not a passive statement saying that a treatment is good. “Should” or “recommend” are good indicators. These may not always be present in an abstract. E.g.:
        \begin{itemize}
            \item “Public policy measures that can reduce inequity in health coverage, as well as improve economic and educational opportunities for the poor, will help in reducing the burden of malaria in SSA.”
        \end{itemize}
        \item \textbf{ETC}: Anything that doesn’t fit into the categories above.
    \end{itemize}
\end{itemize}
All sentences appear in the context of their review. Some of the selected reviews might not actually be reviews; these were identified by accident. These should be excluded from annotation - either make a comment on the side (preferred) or delete the rows belonging to the non-review.

Examples follow. Please ask questions - these guidelines are likely not perfect and we’ll have missed many edge cases

Examples:

\textbf{BACKGROUND}
A sizeable number of individuals who participate in population-based colorectal cancer (CRC) screening programs and have a positive fecal occult blood test (FOBT) do not have an identifiable lesion found at colonoscopy to account for their positive FOBT screen.

\textbf{GOAL}
To determine the effect of integrating informal caregivers into discharge planning on postdischarge cost and resource use in older adults.

\textbf{METHODS}
MAIN OUTCOMES Clinical status (eg, spirometric measures); functional status (eg, days lost from school); and health services use (eg, hospital admissions).
Studies were included if they had measured serum vitamin D levels or vitamin D intake and any periodontal parameter.

\textbf{DETAILED\_FINDINGS}
Overall, 27 studies were included (13 cross-sectional studies, 6 case-control studies, 5 cohort studies, 2 randomized clinical trials and 1 case series study).
Sixty-five percent of the cross-sectional studies reported significant associations between low vitamin D levels and poor periodontal parameters.
Analysis of group cognitive-behavioural therapy (CBT) v. usual care alone (14 studies) showed a significant effect in favour of group CBT immediately post-treatment (standardised mean difference (SMD) -0.55 (95\% CI -0.78 to -0.32)).

\textbf{EFFECT}
This review identified short-term benefits of technology-supported self-guided interventions on the physical activity level and fatigue and some benefit on dietary behaviour and HRQoL in people with cancer.
However, current literature demonstrates a lack of evidence for long-term benefit.

\textbf{EVIDENCE\_QUALITY}
Interpretation of findings was influenced by inadequate reporting of intervention description and compliance.

No meta-analysis was performed due to high variability across studies.

\textbf{RECOMMENDATION}
The decision to perform EGD should be individualized and based on clinical judgement.

\textbf{ETC}
PROSPERO CRD42017080346; \url{https://www.crd.york.ac.uk/prospero/display_record.php?RecordID=80346}.

\subsection{Detailed Sentence Breakdown Scores}

Sentence classification scores for 9 classes are given in Table~\ref{tab:detailed_sentence_breakdown_scores}. The corresponding confusion matrix can be found in Table~\ref{tab:senteces_confusion_matrix}.

\subsection{Sentence Classification Results}
\label{app:sentence_classification_example}

Table \ref{tab:sent_example_full} provides an example of sentence classification results over 3 classes.

\section{Dataset Contradiction Scores}
\label{sec:dataset_confusion_matrix}

The confusion matrix between review effect findings and input study effect findings is given in Table~\ref{tab:review_dataset_confusion_matrix}.

\begin{table*}[t!]
\small
    \centering
    \begin{tabular}{c|cccc}
    \toprule
         & \decreases & \nochange & \increases & count \\
    \midrule 
         \decreases & .338  &.540 & .122 & 1202991\\
         \nochange & .144 & .659 & .197 & 5375546 \\
         \increases & .096 & .529 & .376 & 2490229 \\
    \bottomrule
    \end{tabular}
    \caption{Confusion matrix between review effect findings and input study effect findings. Each row corresponds to the fraction of the effect direction found in the review with the fraction of that direction accounted for in the study. The most frequent confusion is with \nochange, as opposed to flipping the overall direction of the finding.}
    \label{tab:review_dataset_confusion_matrix}
\end{table*}

\begin{table*}[t!]
    \scriptsize
    \centering
    \begin{tabular}{p{55mm}p{46mm}p{46mm}}
        \toprule
        \small{\background} & \small{\target} & \small{\textsc{generated}} \\
        \midrule
        OBJECTIVE To explore the evidence for the effectiveness of acupuncture for nonspecific low back pain ( LBP ). SUMMARY OF BACKGROUND DATA Since the most recent systematic review s on RCTs on acupuncture for LBP, 6 RCTs have been published, which may impact on the previous conclusions. & There is moderate evidence that acupuncture is more  effective  than no treatment, and strong evidence of no significant difference between  acupuncture  and sham acupuncture , for  short-term pain relief. & The is insufficient evidence to support the use of  acupuncture  for LBP. CONCLUSIONS There is limited evidence for the  effectiveness  of acupuncture  in  LBP  in the short term. \\
        \midrule
        Objectives : To provide a quantitative analysis of all r and omized controlled trials design ed to determine the effectiveness of physical interventions for people with spinal cord injury ( SCI ). & There is initial evidence supporting the effectiveness of some physical interventions  for people with SCI. & The Results : This systematic review provides evidence that physical interventions for people with SCI are effective  in  improving  muscle strength and function  in the short term. \\
        \midrule
        BACKGROUND Neuroendocrine tumours ( NET ) most commonly metastasize to the liver. Hepatic resection of NET hepatic metastases ( NETHM ) has been shown to improve symptomology and survival. METHODS A systematic review of clinical studies before September 2010 was performed to examine the efficacy of hepatic resection for NETHM. & Poor histologic grade, extra-hepatic disease and a macroscopically  incomplete resection  were associated with a poor prognosis. CONCLUSION  Hepatic resection  for NETHM provides  symptomatic benefit  and is associated with  favourable survival outcomes  although the majority of patients invariably develop disease progression & Theatic resection of NETHM has been shown to improve  survival  in  patients with advanced, well-differentiated NETs. \\
        \midrule
        The aim of this systematic review and meta- analysis was to assess the efficacy on an intervention on breastfeeding self-efficacy and perceived insufficient milk supply outcomes. & Although significant effect of the interventions in  improving maternal breastfeeding self-efficacy  was revealed by this review, there is still a paucity of evidence on the mode, format, and intensity of interventions. & The findings of this systematic review and meta- analysis suggest that breastfeeding education is an effective intervention for improving  breastfeeding self-efficacy and breastfeeding duration  among  primiparous women. \\
        \bottomrule
    \end{tabular}
    \caption{Example summaries from the test set generated using the \standard configuration.}
    \label{tab:examples}
\end{table*}

\section{Hyperparameters and Modeling Details}
\label{sec:hyperparameters}
We implement our models using PyTorch \citep{pytorch}, the HuggingFace Transformers \citep{huggingface} and PyTorch lightning \citep{lightning} libraries, starting from the BART-base checkpoint \cite{BART}. All models were trained using FP16, using NVidia RTX 8000 GPUs (GPUs with 40G or more of memory are required for most \texttotext configurations). All models are trained for eight epochs as validation scores diminished over time; early experiments ran out to approximately fifty epochs and showed little sensitivity to other hyperparameters. We use gradient accumulation to reach an effective batch size of 32. We use the Adam optimizer \citep{Kingma2015AdamAM} with a learning rate of 1e-5, an epsilon of 1e-8, and a linear learning rate schedule with 1000 steps of warmup. We ran a hyperparameter sweep over decoding parameters on the validation set for 4, 6, and 8 beams; maximum lengths of 64, 128, and 256 wordpieces; and length penalties of 1, 2, and 4. We find little qualitative or quantitative variation between runs and select the setting with the highest Rouge1 scores: 6 beams, a length penalty of 2, and 128 tokens for output maximum lengths. We use an attention dropout \citep{Srivastava2014DropoutAS} of 0.1. Optimizer hyperparameters, as well as any hyperparameters not mentioned, used defaults corresponding to their libraries. Training requires approximately one day on two GPUs. Due to memory constraints, we limit each review to 25 input documents, with a maximum of 1000 tokens per input document. 

We make use of NumPy \citep{2020NumPy-Array} in our models and evaluation, as well as scikit-learn \citep{sklearn_api}, and the general SciPy framework \citep{2020SciPy-NMeth} for evaluation.

\section{Example generated summaries}
\label{appendix:examples}
See Table \ref{tab:examples} for examples of inputs, targets, and generations.

\section{Validation Results}
\label{sec:validation_results}

\begin{table}[t!]
    \centering
    \small
    \begin{tabular}{llllll}
    \toprule
    Model         & R-1     & R-2   & R-L   & $\Delta$EI & F1 \\
    \midrule
    \standard     & 26.66   & 9.04  & 19.78 & .447 & 49.68 \\
    \longbart     & 25.82   & 8.44  & 19.29 & .482 & 47.09 \\
    \bottomrule
    \end{tabular}
    \caption{\texttotext results on the validation set. We report ROUGE, $\Delta$EI, and macro-averaged F1-scores. These are similar to test scores.
    }
    \label{tab:text_to_text_results_validation}
\end{table}

We provide results on the validation set in Tables \ref{tab:text_to_text_results_validation} and \ref{tab:table_to_table_results_validation}.
  
\begin{table}[t!]
    \centering
    \begin{tabular}{llll}
    \toprule
    Model       & P     & R     & F1 \\
    \midrule
    \standard   & 46.98 & 45.39 & 46.97 \\
    \bottomrule
    \end{tabular}
    \caption{\tabtotab results on the validation set. We report precision, recall, and macro-averaged F1-scores. 
    }
    \label{tab:table_to_table_results_validation}
\end{table}

\section{A Brief Review of Systematic Reviews}
\label{sec:overview}

We provide a brief overview of the systematic review process for the reader.
A systematic review is a thorough, evidence-based process to answer scientific questions. 
In the biomedical domain, a systematic review typically consists of five steps: defining the question, finding relevant studies, determining study quality, assessing the evidence (quantitative or qualitative analysis), and drawing final conclusions. 
For a detailed overview of the steps, see \citet{Khan2003FiveST}. While there are other definitions and aspects of the review process \cite{jbievidencesynthesis,Higgins2019CochraneHF}, the five-step process above is sufficient for describing reviews in the context of this work. We emphasize that this work, indeed the approaches used in this field, \emph{cannot} replace the labor done in a systematic review, and may instead be useful for scoping or exploratory reviews. 

The National Toxicology Program,\footnote{\href{https://ntp.niehs.nih.gov/}{https://ntp.niehs.nih.gov/}} part of the United States Department of Health and Human Services, conducts scoping reviews for epidemiological studies. 
The National Toxicology Program has actively solicited help from the natural language processing community via the Text Analysis Conference.\footnote{\href{https://tac.nist.gov/2018/SRIE/}{https://tac.nist.gov/2018/SRIE/}}
Other groups conducting biomedical systematic reviews include the Cochrane Collaboration,\footnote{\href{https://www.cochrane.org/}{https://www.cochrane.org/}} the Joanna Briggs Institute,\footnote{\href{https://jbi.global/}{https://jbi.global/}} Guidelines International Network,\footnote{\href{https://www.g-i-n.net}{https://www.g-i-n.net}} SickKids,\footnote{\href{https://www.sickkids.ca/}{https://www.sickkids.ca/}} the University of York,\footnote{\href{https://www.york.ac.uk/crd/}{https://www.york.ac.uk/crd/}} and the public health agencies of various countries,\footnote{\href{https://www.canada.ca/en/public-health/services/reports-publications.html}{https://www.canada.ca/en/public-health/services/reports-publications.html}} to name a few. Systematic review methodologies have also been applied in fields outside of medicine, by organizations such as the Campbell Collaboration,\footnote{\href{https://www.campbellcollaboration.org}{https://www.campbellcollaboration.org}} which conducts reviews over a wide range of areas: business, justice, education, and more. 

\subsection{Automation in Systematic Reviews}
\label{sec:automation}

Automation in systematic reviews has typically focused on assisting in portions of the process: search and extraction, quality assessment, and interpreting findings. 
For a detailed analysis of automated approaches in aiding the systematic review process, see \citet{norman2020systematic,Marshall2019TowardSR}. 

\textbf{Search and Extraction}. Search, screening, and extracting the results of studies into a structured representation are several components of the systematic review process that have been the major focuses of natural language processing approaches. Several systems provide active-learning enhanced search \cite{sciomescreener,Schoot2020ASReviewOS}, or offer screening based on study type \cite{Marshall2016RobotReviewerEO}.
PICO (\textbf{P}articipants, \textbf{I}nterventions, \textbf{C}ontrols, and \textbf{O}utcomes) elements can be used to assist in search and screening \cite{Znaidi2015AnsweringPC,Ho2016DevelopmentOA,boudin-etal-2010-clinical}. To this end, several datasets have been introduced.
EBM-NLP \cite{ebmnlp} is a dataset of crowd-sourced PICO elements in randomized control trial abstracts. \citet{jin-szolovits-2018-pico} provides a large-scale dataset of sentence-level PICO labels that are automatically derived using the structured abstract headers in PubMed abstracts. The Chemical-Disease Relations challenge \cite{wei2015overview} offers data for some of the PICO classes and a related relation extraction task, as does the i2b2 2010 disease-relation task \cite{i2b22010}.
Evidence Inference \cite{lehman-etal-2019-inferring,evidenceinference} attempts to automate detecting the direction of conclusions given PICO elements of interest; e.g., \citet{nye2020understanding} starts from RCTs, finds PICO elements, and then finds conclusions associated with those PICO elements.
Many review tools\footnote{\href{https://www.evidencepartners.com/}{https://www.evidencepartners.com/}}$^,$\footnote{\href{https://www.covidence.org/reviewers/}{https://www.covidence.org/reviewers/}}$^,$\footnote{\href{https://sysrev.com/}{https://sysrev.com/}}$^,$\footnote{\href{https://www.jbisumari.org/}{https://www.jbisumari.org/}} incorporate workflow management tools for manual extraction of these elements and associated conclusions.

\textbf{Quality Assessment}. Relatively few tools focus on quality assessment. 
The primary tool seems to be RobotReviewer \cite{Marshall2016RobotReviewerEO}, which assesses Risk of Bias in trial results, which is one aspect of quality. There are opportunities for quality assessment that focus on automatically assessing statistical power or study design.

\textbf{Interpretation}. The interpretation step of the systematic review process involves drawing overall conclusions about the interventions studied: how effective is the intervention, when should it be used, what is the overall strength of the evidence supporting the effectiveness and recommendations, and what else needs to be studied. 
It too has received relatively little attention from those developing assistive systems.
Similar to this work, \citet{Wallace2020GeneratingN} takes advantage of structured Cochrane reviews to identify summary targets, and uses portions of the input documents as model inputs.
\citet{shah2021nutribullets} extracts relations from nutritional literature, and uses content planning methods to generate summaries highlighting contradictions in the relevant literature.